\pdfoutput=1

\documentclass[11pt]{article}

\usepackage[]{acl}

\usepackage{times}
\usepackage{latexsym}

\usepackage[T1]{fontenc}

\usepackage[utf8]{inputenc}

\usepackage{microtype}

\usepackage{subfigure}
\usepackage{graphicx}
\usepackage{multirow}
\usepackage{amsmath}
\usepackage{amssymb}

\title{Sentiment Word Aware Multimodal Refinement for Multimodal Sentiment Analysis with ASR Errors}

\author{
	\bf Yang Wu$^1$ ~ Yanyan Zhao$^1$\thanks{~~~Corresponding Author} ~~ Hao Yang$^1$ ~ Song Chen$^1$ ~ Bing Qin$^{1}$\\ \bf ~ Xiaohuan Cao$^{2}$ ~ Wenting Zhao$^{2}$\\
	$^1$ Harbin Institute of Technology  \quad \quad $^2$ AI Lab of China Merchants Bank \\
	$^1$ \{\tt ywu, yyzhao, hyang, songchen, qinb\}@ir.hit.edu.cn \\
	$^2$ \{\tt xhcao, wtzhao\}@cmbchina.com \\
}

\begin{document}
\maketitle
\begin{abstract}
Multimodal sentiment analysis has attracted increasing attention and lots of models have been proposed. However, the performance of the state-of-the-art models decreases sharply when they are deployed in the real world. We find that the main reason is that real-world applications can only access the text outputs by the automatic speech recognition (ASR) models, which may be with errors because of the limitation of model capacity. Through further analysis of the ASR outputs, we find that in some cases the sentiment words, the key sentiment elements in the textual modality, are recognized as other words, which makes the sentiment of the text change and hurts the performance of multimodal sentiment models directly. To address this problem, we propose the sentiment word aware multimodal refinement model (SWRM), which can dynamically refine the erroneous sentiment words by leveraging multimodal sentiment clues. Specifically, we first use the sentiment word position detection module to obtain the most possible position of the sentiment word in the text and then utilize the multimodal sentiment word refinement module to dynamically refine the sentiment word embeddings. The refined embeddings are taken as the textual inputs of the multimodal feature fusion module to predict the sentiment labels. We conduct extensive experiments on the real-world datasets including MOSI-Speechbrain, MOSI-IBM, and MOSI-iFlytek and the results demonstrate the effectiveness of our model, which surpasses the current state-of-the-art models on three datasets. Furthermore, our approach can be adapted for other multimodal feature fusion models easily. 

\end{abstract}

\section{Introduction}
Multimodal sentiment analysis (MSA) has been an emerging research field for its potential applications in human-computer interaction. How to effectively fuse multimodal information including textual, acoustic, and visual to predict the sentiment is a very challenging problem and has been addressed by many previous studies. Some works focus on introducing additional information into the fusing model, such as the alignment information between different modal features \cite{wu-etal-2021-text} and unimodal sentiment labels \cite{Yu_Xu_Yuan_Wu_2021}. And other works consider the semantic gaps between multimodal data and adopt the adversarial learning \cite{Mai_Hu_Xing_2020} and multi-task learning \cite{10.1145/3394171.3413678} to map different modal features into a shared subspace. 

\begin{figure}[t]
	\begin{center}
		\subfigure[Results of the Self-MM model on the real-world datasets. SpeechBrain, IBM, and iFlytek are three ASR APIs we adopted.]{
			\includegraphics[width=0.6\linewidth]{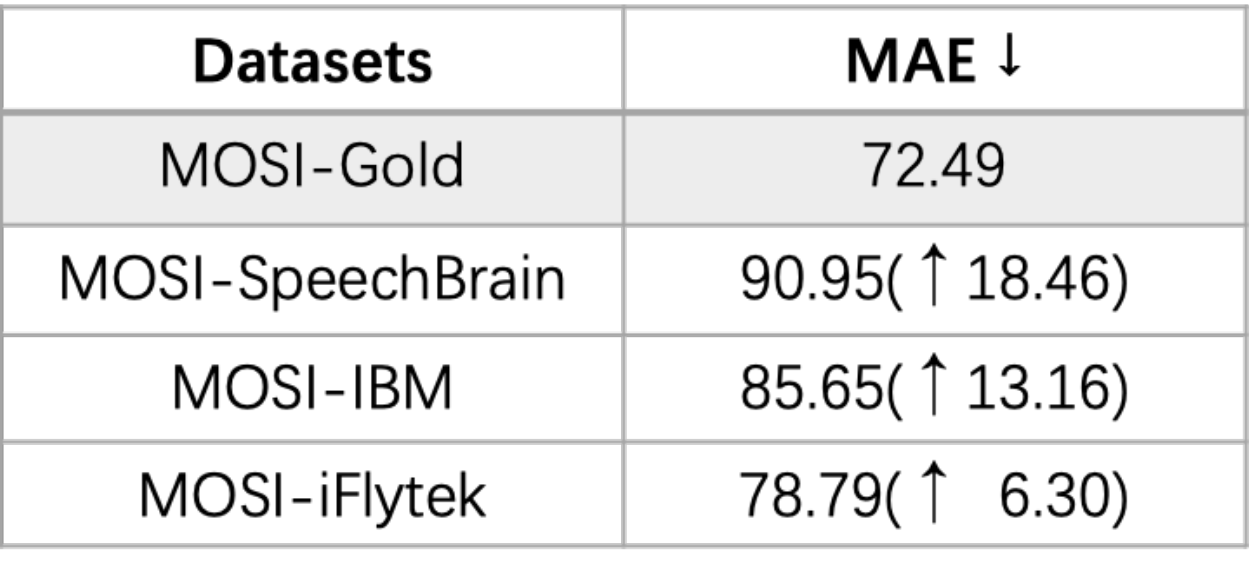}
			\label{fig1a}
		}
		\subfigure[An example of the sentiment word substitution error and the percentages of it on the datasets.]{
			\includegraphics[width=0.7\linewidth]{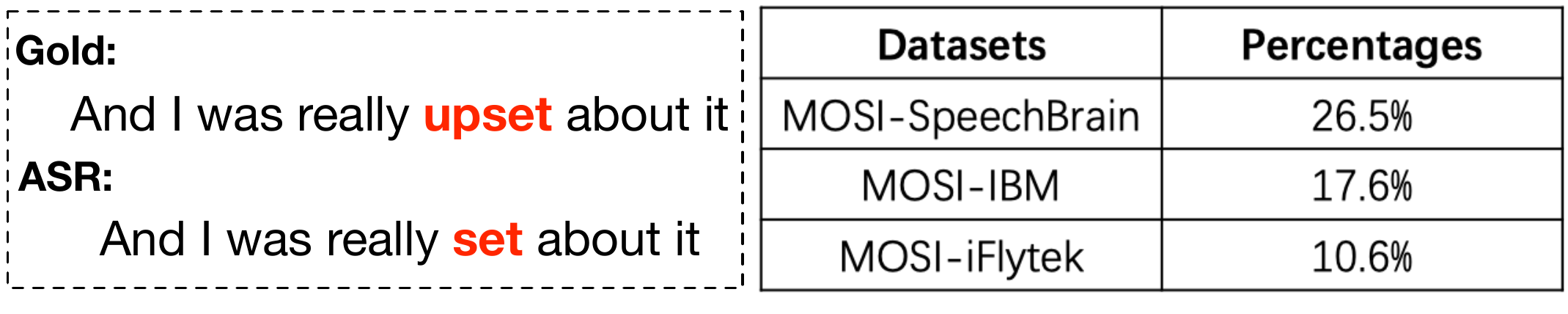}
			 \label{fig1b}
		}	
		\subfigure[Our approach to reduce the negative impact of the sentiment word substitution error on the MSA models.]{
			\includegraphics[width=0.7\linewidth]{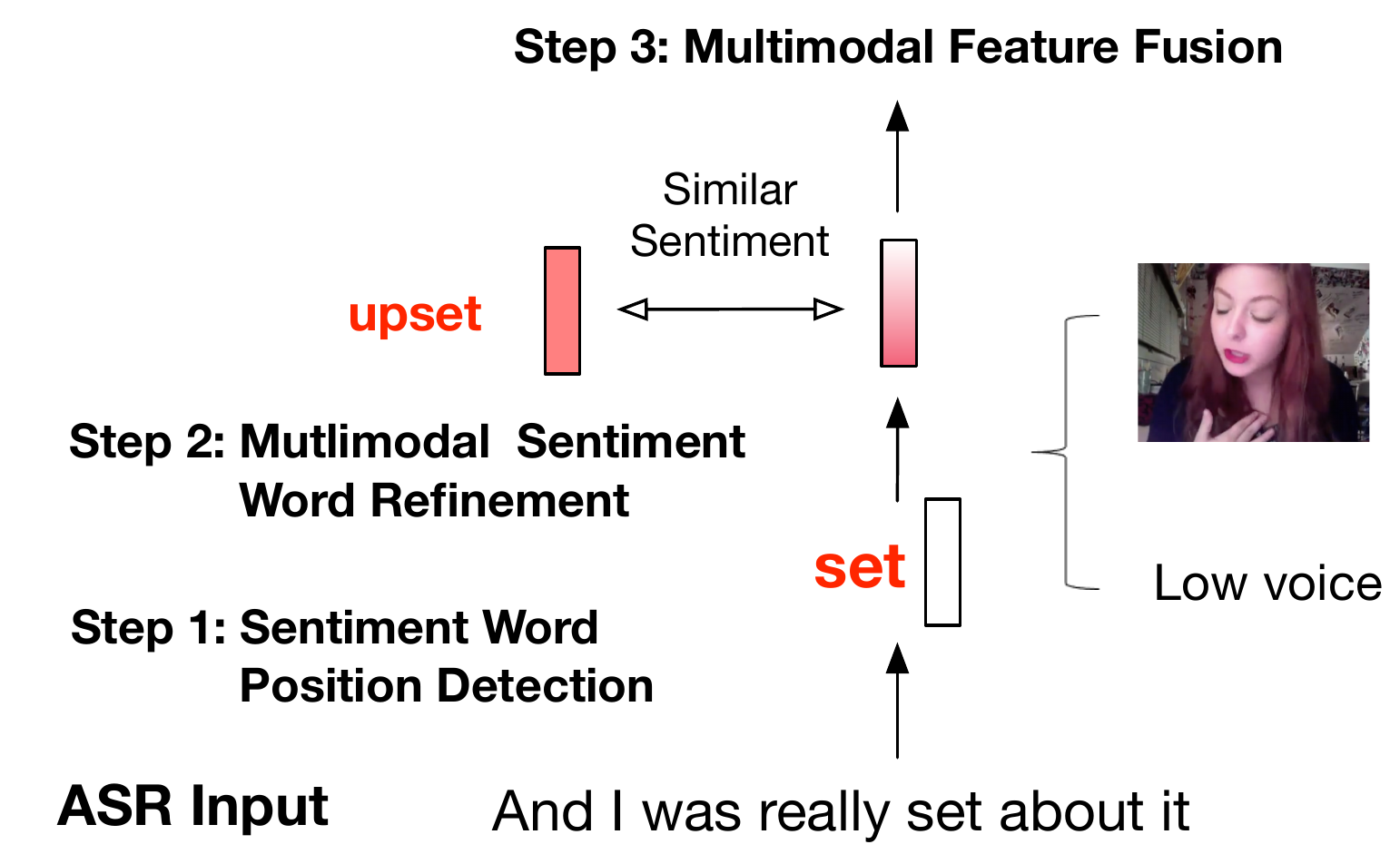}
			 \label{fig1c}
		}		
		
	\end{center}
	\caption{Illustration of our motivation. }
\end{figure}

Despite the apparent success of the current state-of-the-art models, their performance decreases sharply, when deployed in the real world. The reason is that the input texts are provided by the ASR models, which usually are with errors because of the limitation of model capacity. To further analyze this problem, we build three real-world multimodal sentiment analysis datasets based on the existing dataset, CMU-MOSI\cite{zadeh2016mosi}. Specifically, we adopt three widely used ASR APIs including SpeechBrain, IBM, and iFlytek to process the original audios and obtain the recognized texts. Then, we replace the gold texts in CMU-MOSI with the ASR results and get three real-world datasets, namely MOSI-SpeechBrain, MOSI-IBM, and MOSI-iFlytek. We evaluate the current state-of-the-art model, Self-MM\cite{Yu_Xu_Yuan_Wu_2021}, and report the mean absolute error (MAE) on the multimodal sentiment analysis task. As we can see in Figure \ref{fig1a}, when the model is deployed in the real world, there is an obvious drop in model performance.

The further in-depth analysis of ASR errors shows that the sentiment word substitution error can hurt the MSA model directly. The reason is that the sentiment words in the text are the most important clues in the textual modality for detecting sentiment and incorrectly recognizing them could change the sentiment conveyed by the text. To have an intuitive understanding of the sentiment word substitution error, we take an example in Figure \ref{fig1b}. The gold text is ``And I was really upset about it", but the ASR model (SpeechBrain) recognizes the sentiment word ``upset" wrongly as ``set", which results in the change of the sentiment semantics of the text and directly affects the MSA model performance. We list the percentages of the sentiment word substitution error on the MOSI dataset for three ASR APIs in Figure \ref{fig1b}. The percentage of the sentiment word substitution error on the MOSI-IBM is 17.6\%, which means about 17 of 100 utterances have this type of error. To further demonstrate the negative effect of the substitution error on the MSA models, we split the test data of MOSI-IBM into two groups by whether there is a substitution error. We evaluate Self-MM on the test data and observe that the misclassification rate of the group in which the substitution error exists is higher than the other group (29.9\% vs 15.8\%). This result indicates that the sentiment word substitution error could hurt the state-of-the-art MSA model.

To tackle this problem, we propose the sentiment word aware multimodal refinement model, which can detect the positions of the sentiment words in the text and dynamically refine the word embeddings in the detected positions by incorporating multimodal clues. The basic idea of our approach is shown in Figure \ref{fig1c}. We consider leveraging the multimodal sentiment information, namely the negative sentiment conveyed by the low voice and sad face, and textual context information to help the model reconstruct the sentiment semantics for the input embeddings. Specifically, we first use the sentiment word location module to detect the positions of sentiment words and meanwhile utilize the strong language model, BERT, to generate the candidate sentiment words. Then we propose the multimodal sentiment word refinement module to refine the word embeddings based on the multimodal context information. The refinement process consists of two parts, filtering and adding. We apply the multimodal gating network to filter out useless information from the input word embeddings in the filtering process and use the multimodal sentiment word attention network to leverage the useful information from candidate sentiment words as the supplement to the filtered word embeddings in the adding process. Finally, the refined sentiment word embeddings are used for multimodal feature fusion.

We conduct extensive experiments on the MOSI-SpeechBrain, MOSI-IBM, and MOSI-iFlytek datasets to demonstrate the effectiveness of our proposed model. The experimental results show that: (1) There is an obvious performance drop for the state-of-the-art MSA model, when the model is deployed in the real world taking the ASR outputs as the input of textual modality; (2) Our proposed model outperforms all baselines, which can dynamically refine the sentiment word embeddings by leveraging multimodal information.

The main contributions of this work are as follows: (1) We propose a novel sentiment word aware multimodal refinement model for multimodal sentiment analysis, which can dynamically reconstruct the sentiment semantics of the ASR texts with errors by utilizing the multimodal sentiment information resulting in more robust sentiment prediction; (2) We validate the negative effect of the sentiment word substitution error on the state-of-the-art MSA model through the in-depth analysis; (3) We evaluate our model on three real-world datasets, and the experimental results demonstrate that our model outperforms all baselines.

\section{Related Work}

Multimodal sentiment analysis has gained increasing attention from the community recently and some process has been made. In general, there are three findings presented by previous work. 

\textbf{Performing the cross-modal alignment is helpful for multimodal feature fusion}. \citet{10.1145/3136755.3136801} considered that the holistic features mainly contain global information, which may fail to capture local information. Therefore, they applied the force-alignment to align the visual and acoustic features with the words and further obtained the word-level features. To effectively fuse them, they proposed the GME-LSTM(A) model, which consists of two modules, the gated multimodal embedding and the LSTM with the temporal attention. However, obtaining the word-level features needs to perform the force-alignment, which is time-consuming. To address it, \citet{tsai-etal-2019-multimodal} proposed the MulT model, which uses the cross-modal attention to align different modal features implicitly. Instead of performing the alignment in the time dimension, some works focusing on semantic alignment. \citet{10.1145/3394171.3413678} considered that the semantic gaps between heterogeneous data could hurt the model performance and proposed the MISA model, which maps the different modal data into a shared space before multimodal feature fusion. \citet{wu-etal-2021-text} first utilized the cross-modal prediction task to distinguish the shared and private semantics of non-textual modalities compared to the textual modality and then fuse them. The above works show that performing the cross-modal alignment is helpful for multimodal feature fusion.

\textbf{Training the MSA models in an end-to-end manner is more effective}. Most of the previous studies adopt a two-phase pipeline, first extracting unimodal features and then fusing them. \citet{dai-etal-2021-multimodal} considered that it may lead to sub-optimal performance since the extracted unimodal features are fixed and cannot be further improved benefiting from the downstream supervisory signals. Therefore, they proposed the multimodal end-to-end sparse model, which can optimize the unimodal feature extraction and multimodal feature fusion jointly. The experimental results on the multimodal emotion detection task show that training the models in an end-to-end manner can obtain better results than the pipeline models.

\textbf{Leveraging the unimodal sentiment labels to learn more informative unimodal representations is useful for multimodal feature fusion}. \citet{yu-etal-2020-ch} considered that introducing the unimodal sentiment labels can help the model capture the unimodal sentiment information and model the difference between modalities. Motivated by it, they built the CH-SIMS dataset, which contains not only the multimodal sentiment labels but also unimodal sentiment labels. And based on it, they proposed a multi-task learning framework to leverage two types of sentiment labels simultaneously. However, this method needs unimodal labels, which is absent for most of the existing datasets. To address it, \citet{Yu_Xu_Yuan_Wu_2021} proposed the Self-MM model, which first generates the unimodal labels by utilizing the relationship between the unimodal and multimodal labels and then uses the multi-task learning to train the model. These two works both address the usefulness of introducing unimodal labels.

However, even though lots of models are proposed and obtain promising results on the benchmark datasets, there are few works considering the noisy inputs when the MSA models are deployed in the real world. \citet{10.1145/3136755.3136801} presented the Gated Multimodal Embedding to filter out the noises from the acoustic and visual data. \citet{Pham_Liang_Manzini_Morency_2019} considered that visual and acoustic data may be absent and proposed the MCTN model to handle it.  \citet{liang-etal-2019-learning} and \citet{Mittal_Bhattacharya_Chandra_Bera_Manocha_2020} also mainly focused on dealing with the noises introduced by the visual and acoustic data, and their models are based on the word-level features, which are obtained by aligning the audios with the gold texts. There is only one work \cite{dumpala2018sentiment} considering that the texts are output by the ASR models, which may be erroneous. But this work does not study how do the ASR errors affect the MSA models and does not evaluate the SOTA MSA models on the datasets. Besides, the proposed model needs the gold texts when training, which is time-consuming and labor-consuming.

Comparing to the above works, we evaluate the SOTA MSA models on the real-world datasets and observe that the performance of models decreases sharply because of the erroneous ASR texts. Through in-depth analysis of the ASR outputs, we find the sentiment word substitution error in the ASR texts could hurt the MSA models directly. To address it, we propose the sentiment word aware multimodal refinement model, which only uses the ASR texts in the training and testing phrases.

\begin{figure*}[t]
\centering
\includegraphics[width=1.7\columnwidth]{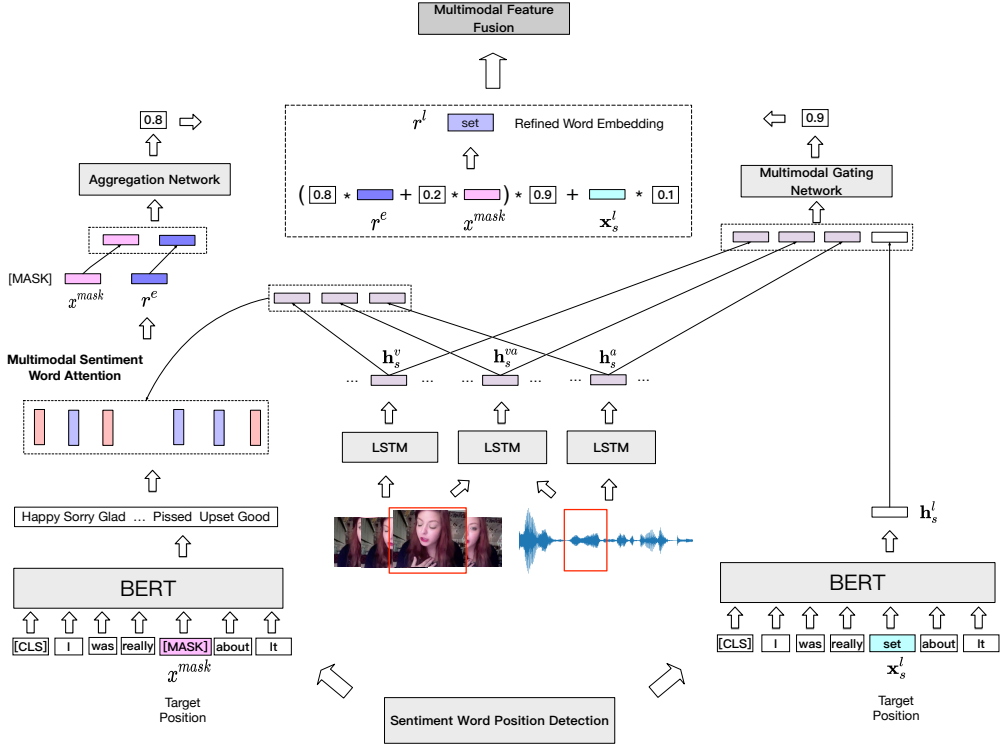} 
\caption{Illustration of our proposed model.}.
\label{model}
\end{figure*}

\section{Approach}

In this section, we describe the sentiment word aware multimodal refinement model in detail. An illustration of our proposed model is given in Figure \ref{model}. Our model consists of three modules including the sentiment word location module, multimodal sentiment word refinement module, and multimodal feature fusion module. We first use the sentiment word location module to detect the possible positions of sentiment words and then utilize the multimodal sentiment word refinement module to dynamically refine the word embeddings in the detected positions. Finally, the refined word embeddings are fed into the multimodal feature fusion module to predict the final sentiment labels.

\subsection{Sentiment Word Position Detection}
The core idea of the sentiment word position detection module is to find out the possible positions of sentiment words in the ASR texts. Note that, it is different from locating sentiment words depending on the word semantics, since the ASR models may recognize a sentiment word as a neutral word, which makes it hard to locate correctly. For example, given a gold text ``And I was really upset about it", the ASR model recognizes it as ``And I was really set about it". It is easy for the model to label the word ``set" as a neutral word. Therefore, we choose to detect the position of the sentiment words instead of locating them. 

To achieve it, we consider adopting a powerful language model, since the language model can model the context information of the sentiment words such as syntactic and grammatical information and predict the appropriate words for the target position. Specifically, we choose the BERT model \cite{devlin-etal-2019-bert} as our language model since the masked language modeling pretraining objective meets our needs perfectly. Given the sentence $\{w_1, w_2, ..., w_{n_l}\}$, we first mask each word $w_i$ in the sentence sequentially, and in practice, we replace the word with the special word [MASK]. For example, we mask the first word in the sentence and obtain $\{{\rm [MASK]}, w_2, ..., w_{n_l}\}$. And then we use the BERT model to predict the possible words in the position of the masked word. We sort the predicted candidate words by the prediction probabilities and get the Top-k candidate words $C_i = \{c_1^i, c_2^i, ... ,c_k^i\}$.   

Next, we distinguish the sentiment words from the candidates using the sentiment lexicons \cite{hu2004mining, wilson2005recognizing} and $k_i$ is the number of selected sentiment words corresponding to the position $i$. The larger the number is, the more possible the position is. And we obtain the most possible position of sentiment word, $s = \mathop{\arg\max}(\{k_1, k_2, ..., k_{n_l}\})$. Considering that in some cases there is not a sentiment word in the sentence, we use a sentiment threshold to filter out the impossible ones. In practice, we use the gate mask $p$ to record it,  and $p$ is 1 if  $k_s$ is larger than $k/2$ and 0 otherwise.

\subsection{Multimodal Sentiment Word Refinement}

In order to reduce the negative effects of the ASR errors, we propose the multimodal sentiment word refinement module, in which we refine the word embeddings of sentiment words from two aspects. One is that we uses the multimodal gating network to filter out the useless information from the input word embeddings. The other one is that we design the multimodal sentiment attention network to incorporate the useful information from candidate words generated by the BERT model.

Given an utterance, which includes three modal unaligned features, word embeddings, acoustic features, and visual features, we denote them as  $\mathbf x^i=\{x_t^i : 1 \le t \le n_i, x_t^i \in \mathbb R^{d_x^i}\} $, $i \in \{l, v, a\}$. To obtain the multimodal information corresponding to each word, We utilize the pseudo-alignment method to align the features. We split the the acoustic and visual features into non-overlapping feature groups, of which lengths are $\left\lfloor \frac{n_a}{n_l} \right\rfloor$ and $\left\lfloor \frac{n_v}{n_l} \right\rfloor$ respectively, and average the features in each group and obtain the aligned features,  $\mathbf u^i=\{u_t^i : 1 \le t \le n_l, u_t^i \in \mathbb R^{d_x^i}\} $, $i \in \{v, a\}$.

To obtain the context-aware representations, we apply the BERT model and LSTM networks to encode the features, producing $\mathbf h^i=\{h_t^i : 1 \le t \le n_l, h_t^i \in \mathbb R^{d^i_h}\}$, $i \in \{v, a, l\}$. Besides, we also use an LSTM network to fuse the acoustic and visual features for capturing high-level sentiment semantics and obtain $\mathbf h^{va}=\{h_t^{va} : 1 \le t \le n_{l}, h_t^{va} \in \mathbb R^{d^{va}_h}\}$. 

\begin{equation}
\begin{aligned}
\mathbf h^l &=\rm {BERT}(\mathbf x^l) \\
\mathbf h^v &=\rm {LSTM_v}(\mathbf u^v) \\
\mathbf h^a &=\rm {LSTM_a}(\mathbf u^a) \\ 
\mathbf h^{va} &=\rm {LSTM_{va}}([\mathbf u^v; \mathbf u^a]) \\
\end{aligned}
\end{equation}

Subsequently, We propose the multimodal gating network to filter the word embedding, which is implemented by a non-linear layer. The motivation is that the ASR model may recognize incorrectly the sentiment word resulting in the corrupted sentiment semantics of the text. Therefore, we leverage the multimodal sentiment information to decide how much information of the input word embedding to pass. Specifically, we concatenate the unimodal context-aware representations, $\mathbf h^l_s$, $\mathbf h^v_s$, $\mathbf h^a_s$, and bimodal representation $\mathbf h^{va}_s$ in the detected position $s$ and feed them into a non-linear neural network, producing the gate value $g^v$. And then the gate value is used to filter out the useless information from the word embedding. To make the model ignore the impossible one, we use the gate mask $p$ to achieve it.

\begin{equation}
\begin{aligned}
g^v &= Sigmoid(W_1([\mathbf h^l_s; \mathbf h^v_s; \mathbf h^a_s; \mathbf h^{va}_s]) + b_1) \\
r^v &=  (1 - g^vp) \mathbf x^l_s
\end{aligned}
\end{equation}

where $W_{1} \in \mathbb R^{1 \times \sum_{i \in \{l, v, a, va\}} d^i_h}$, $b_1 \in \mathbb R^1$ are the parameters of the multimodal gating network.

Furthermore, we propose a novel multimodal sentiment word attention network to leverage the sentiment-related information from the candidate words, more than half of which are sentiment words, generated by the BERT model to complement the word embeddings. For example, the ASR model recognizes the ``upset" as ``set", we first want to remove the useless information of ``set" and then incorporate the information of negative sentiment words to reconstruct the original sentiment semantics. We use a linear layer to implement the multimodal sentiment word attention network. We first concatenate the word embedding $x^{c_t^s}$ of the candidate word $c_t^s$ and multimodal representations, $\mathbf h_{s}^v$, $\mathbf h_{s}^a$, and $\mathbf h_{s}^{va}$ at the most possible time step $s$. Then, we pass them to the linear layer and obtain the attention score $g_t^e$. The attention scores are fed into a softmax function to obtain the attention weights. Finally, we apply the weights to the candidate word embeddings and get the sentiment embedding $r^e$.

\begin{equation}
\begin{aligned}
g_t^e &= W_2([  x^{c_t^s}; \mathbf h_{s}^v; \mathbf h_{s}^a; \mathbf h_{s}^{va}]) + b_2 \\
w_{t}^{e} &=  \frac{e^{g_{t}^{e}}}{\sum_{t=1}^{k}{e^{g_{t}^{e}}}} \\
r^e &= \sum_{t=1}^{k} w_t^{e}x^{c_t^s} \\
\end{aligned}
\end{equation}

where $W_{2} \in \mathbb R^{1 \times (d_x^l + \sum_{i \in \{v, a, va\}} d^i_h)}$, $b_2 \in \mathbb R^1$ are the parameters of the multimodal sentiment word attention network.

In addition, there may not be suitable words in the candidate words. Hence, we incorporate the embedding of the special word $[\rm MASK]$, $x^{mask}$, to let the BERT model handle this problem based on the context. We then design an aggregation network to balance the contributions of the special word embedding $x^{mask}$ and the sentiment embedding $r^e$. Finally, we add the $r^{add}$ to the filtered word embedding $u_s^l$ and obtain the refined word embedding $r_l$ for the target word.  

\begin{equation}
\begin{aligned}
g^{mask} &= Sigmoid(W_3([r^e; x^{mask}]) + b_3) \\
r^{add} &= g^{mask} r^e + (1-g^{mask}) x^{mask} \\
 r^l &= (g^v p) r^{add} + r^v
\end{aligned}
\end{equation}

where $W_{3} \in \mathbb R^{1 \times 2d_x^l} $, $b_3 \in \mathbb R^1$ are the trainable parameters.

\subsection{Multimodal Feature Fusion}

We describe our multimodal feature fusion module in the section and it is noted that our proposed refinement approach only modifies the textual input token embeddings, which makes it easy to be adapted for other multimodal feature fusion models based on BERT, such as MISA \cite{10.1145/3394171.3413678}.

We first use the BERT model to encode the refined word embeddings $\mathbf{z^l} = \{x^l_1, x^l_2, ..., r^l, .., x^l_{n_l}\}$ and take the representation of [CLS] as the textual representation, which is denoted as $v^l$. And then we use two LSTM networks to encode the visual and acoustic features and take the representations of the first words as the visual representation $v^v$ and acoustic representation $v^a$. Finally, we fuse them using a non-linear layer to capture the interactions between them. 

\begin{equation}
\begin{aligned}
v^l &=\rm {BERT_{textual}}(\mathbf z^l) \\
v^v &=\rm {LSTM_{visual}}(\mathbf x^v) \\
v^a &=\rm {LSTM_{acoustic}}(\mathbf x^a) \\
v^f &= Relu(W_4([v^l;v^v;v^a])+b_4)
\end{aligned}
\end{equation}

where $W_{4} \in \mathbb R^{d_v^f \times (d_v^l + d_v^a + d_v^v)} $, $b_4 \in \mathbb R^{d_v^f}$ are the trainable parameters of the fusion network.

We utilize a linear layer to predict the final sentiment regression labels.
\begin{equation}
\begin{aligned}
p_f &= W_5v^f+b_5
\end{aligned}
\end{equation}

where $W_{5} \in \mathbb R^{1 \times d_v^f} $, $b_5 \in \mathbb R^1$ are the trainable parameters of the prediction network.

Besides, to enhance the model to capture unimodal sentiment information,  we use the Unimodal Label Generation Module (ULGM) \cite{Yu_Xu_Yuan_Wu_2021} to generate pseudo unimodal sentiment labels and adopt them to train our model in a multi-task learning manner. For more details, we refer you to \citet{Yu_Xu_Yuan_Wu_2021}.


\begin{table*}[tbp]
	\centering
	\resizebox{\linewidth}{!}{
	\begin{tabular}{l|l|llllll}
		\hline
		\multicolumn{1}{c|}{\multirow{2}{*}{Datasets}}  & \multicolumn{1}{c|}{\multirow{2}{*}{Models}}                & \multicolumn{6}{c}{Evaluation Metrics}                           \\
         &  & \multicolumn{1}{c}{Has0-Acc $\uparrow$} & \multicolumn{1}{c}{Has0-F1 $\uparrow$} &  \multicolumn{1}{c}{Non0-Acc $\uparrow$} & \multicolumn{1}{c}{Non0-F1 $\uparrow$} & \multicolumn{1}{c}{MAE $\downarrow$}       & \multicolumn{1}{c}{Corr $\uparrow$} \\
		\hline\hline
        \multicolumn{1}{c|}{\multirow{6}{*}{MOSI-SpeechBrain}}  & \multicolumn{1}{c|}{TFN(B)} & \multicolumn{1}{c}{68.98} & \multicolumn{1}{c}{68.95} &  \multicolumn{1}{c}{69.51} & \multicolumn{1}{c}{69.57} & \multicolumn{1}{c}{115.55}       & \multicolumn{1}{c}{48.54} \\
        & \multicolumn{1}{c|}{LMF(B)} & \multicolumn{1}{c}{68.86} & \multicolumn{1}{c}{68.88} &  \multicolumn{1}{c}{69.36} & \multicolumn{1}{c}{69.48} & \multicolumn{1}{c}{117.42}       & \multicolumn{1}{c}{48.66} \\
		  & \multicolumn{1}{c|}{MulT(B)} & \multicolumn{1}{c}{71.78} & \multicolumn{1}{c}{71.70} &  \multicolumn{1}{c}{72.74} & \multicolumn{1}{c}{72.75} & \multicolumn{1}{c}{109.00}       & \multicolumn{1}{c}{54.69} \\
		  & \multicolumn{1}{c|}{MISA} & \multicolumn{1}{c}{73.79} & \multicolumn{1}{c}{73.85} &  \multicolumn{1}{c}{74.51} & \multicolumn{1}{c}{74.66} & \multicolumn{1}{c}{98.52}       & \multicolumn{1}{c}{65.37} \\
		  & \multicolumn{1}{c|}{Self-MM} & \multicolumn{1}{c}{73.67} & \multicolumn{1}{c}{73.72} &  \multicolumn{1}{c}{74.85} & \multicolumn{1}{c}{74.98} & \multicolumn{1}{c}{90.95}       & \multicolumn{1}{c}{67.23} \\
		  \cline{2-8}
		  & \multicolumn{1}{c|}{Ours} & \multicolumn{1}{c}{\textbf{74.58}} & \multicolumn{1}{c}{\textbf{74.62}} &  \multicolumn{1}{c}{\textbf{75.70}} & \multicolumn{1}{c}{\textbf{75.82}} & \multicolumn{1}{c}{\textbf{90.56}}       & \multicolumn{1}{c}{\textbf{67.47}} \\
		  \hline
        \multicolumn{1}{c|}{\multirow{6}{*}{MOSI-IBM}}   & \multicolumn{1}{c|}{TFN(B)} & \multicolumn{1}{c}{71.81} & \multicolumn{1}{c}{71.78} &  \multicolumn{1}{c}{72.13} & \multicolumn{1}{c}{73.21} & \multicolumn{1}{c}{109.42}       & \multicolumn{1}{c}{58.19} \\
        & \multicolumn{1}{c|}{LMF(B)} & \multicolumn{1}{c}{73.06} & \multicolumn{1}{c}{73.09} &  \multicolumn{1}{c}{74.30} & \multicolumn{1}{c}{74.41} & \multicolumn{1}{c}{104.70}       & \multicolumn{1}{c}{59.07} \\
		  & \multicolumn{1}{c|}{MulT(B)} & \multicolumn{1}{c}{75.57} & \multicolumn{1}{c}{75.54} &  \multicolumn{1}{c}{76.74} & \multicolumn{1}{c}{76.79} & \multicolumn{1}{c}{100.32}       & \multicolumn{1}{c}{64.34} \\
		  & \multicolumn{1}{c|}{MISA} & \multicolumn{1}{c}{76.97} & \multicolumn{1}{c}{76.99} &  \multicolumn{1}{c}{78.08} & \multicolumn{1}{c}{78.17} & \multicolumn{1}{c}{91.23}       & \multicolumn{1}{c}{71.30} \\
		  & \multicolumn{1}{c|}{Self-MM} & \multicolumn{1}{c}{77.32} & \multicolumn{1}{c}{77.37} &  \multicolumn{1}{c}{78.60} & \multicolumn{1}{c}{78.72} & \multicolumn{1}{c}{85.65}       & \multicolumn{1}{c}{73.23} \\
		  \cline{2-8}
		  & \multicolumn{1}{c|}{Ours} & \multicolumn{1}{c}{\textbf{78.43}} & \multicolumn{1}{c}{\textbf{78.47}} &  \multicolumn{1}{c}{\textbf{79.70}} & \multicolumn{1}{c}{\textbf{79.80}} & \multicolumn{1}{c}{\textbf{82.91}}       & \multicolumn{1}{c}{\textbf{73.91}} \\
		  \hline
		  \multicolumn{1}{c|}{\multirow{6}{*}{MOSI-iFlytek}}   & \multicolumn{1}{c|}{TFN(B)} & \multicolumn{1}{c}{71.95} & \multicolumn{1}{c}{72.01} &  \multicolumn{1}{c}{72.62} & \multicolumn{1}{c}{72.76} & \multicolumn{1}{c}{107.01}       & \multicolumn{1}{c}{56.52} \\
		  & \multicolumn{1}{c|}{LMF(B)} & \multicolumn{1}{c}{71.98} & \multicolumn{1}{c}{72.03} &  \multicolumn{1}{c}{72.35} & \multicolumn{1}{c}{72.49} & \multicolumn{1}{c}{106.63}       & \multicolumn{1}{c}{59.48} \\
		  & \multicolumn{1}{c|}{MulT(B)} & \multicolumn{1}{c}{77.32} & \multicolumn{1}{c}{77.05} &  \multicolumn{1}{c}{78.75} & \multicolumn{1}{c}{78.56} & \multicolumn{1}{c}{89.84}       & \multicolumn{1}{c}{68.14} \\
		  & \multicolumn{1}{c|}{MISA} & \multicolumn{1}{c}{79.59} & \multicolumn{1}{c}{79.62} &  \multicolumn{1}{c}{79.82} & \multicolumn{1}{c}{79.91} & \multicolumn{1}{c}{85.63}       & \multicolumn{1}{c}{74.53} \\
		  & \multicolumn{1}{c|}{Self-MM} & \multicolumn{1}{c}{80.26} & \multicolumn{1}{c}{80.26} &  \multicolumn{1}{c}{81.16} & \multicolumn{1}{c}{81.20} & \multicolumn{1}{c}{78.79}       & \multicolumn{1}{c}{75.83} \\
		  \cline{2-8}
		  & \multicolumn{1}{c|}{Ours} & \multicolumn{1}{c}{\textbf{80.47}} & \multicolumn{1}{c}{\textbf{80.47}} &  \multicolumn{1}{c}{\textbf{81.28}} & \multicolumn{1}{c}{\textbf{81.34}} & \multicolumn{1}{c}{\textbf{78.39}}       & \multicolumn{1}{c}{\textbf{75.97}} \\
		  \hline
		  \multicolumn{1}{c|}{MOSI-Gold}  & \multicolumn{1}{c|}{Self-MM} & \multicolumn{1}{c}{82.54} & \multicolumn{1}{c}{82.51} &  \multicolumn{1}{c}{84.02} & \multicolumn{1}{c}{84.05} & \multicolumn{1}{c}{72.49}       & \multicolumn{1}{c}{78.90} \\
		  \hline
        \end{tabular}
    }
    \caption{Results on the MOSI-SpeechBrain, MOSI-IBM, and MOSI-iFlytek datasets. (B) means the textual features are based on BERT. The best results are in \textbf{bold}.}
	\label{tab:main_mosi}
\end{table*}

\section{Experiment}

\subsection{Datasets}
We build three real-world datasets including MOSI-SpeechBrain, MOSI-IBM, and MOSI-iFlytek, on CMU-MOSI\cite{zadeh2016mosi}. 

\paragraph{CMU-MOSI} CMU multimodal opinion-level sentiment intensity (CMU-MOSI) consists of 93 videos collected from the YouTube website. The length of the videos varies from 2-5 mins. These videos are split into 2,199 short video clips and labeled with sentiment scores from -3 (strongly negative) to 3 (strongly positive). For multimodal features, we extract the visual features using Facet, which can extract the facial action units \cite{ekman1980facial} from each frame. The acoustic features are obtained by applying COVAREP \cite{degottex2014covarep}, which includes 12 Mel-frequency cepstral coefficients (MFCCs) and other low-level features. 

However, the provided texts of the utterances in the MOSI dataset are manually transcribed from the corresponding videos by the expert transcribers, which is unrealistic for the real-world applications to obtain the texts in such a way. To evaluate the models in the real world, we replace the manually gold texts in the dataset with the texts output by the ASR models. We adopt a strong ASR model and two widely used commercial APIs to produce the texts. The utilized ASR model released by \citet{speechbrain} is built on the transformer encoder-decoder framework and trained on the Librispeech dataset\cite{panayotov2015librispeech}. The commercial APIs used by us are IBM\footnote{https://www.ibm.com/cloud/watson-speech-to-text} and iFlytek\footnote{https://global.xfyun.cn/products/lfasr} speech-to-text APIs, which are wildly used by researchers and software developers. Finally, we apply the three ASR models to transcribe the videos into texts and construct three new datasets, namely MOSI-SpeechBrain, MOSI-IBM, and MOSI-iFlytek. We report the WER results of the adopted ASR models on MOSI in Appendix \ref{appdendix_a}. Noted that, we do not adopt MOSEI \cite{bagher-zadeh-etal-2018-multimodal}, because it does not provide the original video clips for the extracted features and annotated sentiment labels, and we can not process the original audios. 

\subsection{Training Details}
We use Adam as the optimizer and the learning rate is 5e-5. The batch size is 64. The sentiment threshold is set to 0.5 while detecting the sentiment word position. The number of the candidate words $k$ is 50. The other hyper-parameters of the model are reported in Appendix \ref{appdendix_b}. All experiments are run on an Nvidia Tesla P100 GPU. We run five times and report the average performance. The random seeds we used are 1111,1112, 1113, 1114, and 1115.

\subsection{Evaluation Metrics}
For the MOSI-SpeechBrain, MOSI-IBM, and MOSI-iFlytek datasets, following previous work \cite{Yu_Xu_Yuan_Wu_2021}, we take 2-class accuracy(Acc-2), F1 score(F1), mean absolute error (MAE), and correlation(Corr) as our evaluation metrics. And for Acc-2 and F1-Score, we calculate them in two ways, negative/non-negative (Non0-Acc, Non0-F1) and negative/positive (Has0-Acc, Has0-F1). As the prediction results are real values, we obtain the sentiment classification labels by mapping the sentiment scores into labels.

\subsection{Baselines}
We compare our proposed model with the following baselines \footnote{Because applying the force-alignment using the errorous ASR texts leads to cascading errors resulting in poor aligned features, we only take the models using unaligned features as our baselines for a fair comparison. }. \textbf{TFN}  \cite{zadeh-etal-2017-tensor} uses the three-fold Cartesian product to capture unimodal, bimodal, and trimodal interactions. \textbf{LMF}  \cite{liu-etal-2018-efficient} uses the low-rank tensors to accelerate the multimodal feature fusion process. \textbf{MulT} \cite{tsai-etal-2019-multimodal} uses the cross-modal transformers to fuse multimodal features. \textbf{MISA} \cite{10.1145/3394171.3413678} adopts multi-task learning to map different modal features into a shared subspace. \textbf{Self-MM} \cite{Yu_Xu_Yuan_Wu_2021} first generates the pseudo unimodal sentiment labels and then adopts them to train the model in a multi-task learning manner.

\section{Results and Analysis}
\subsection{Quantitative Results}
In Table \ref{tab:main_mosi}, we show the results on the MOSI-SpeechBrain, MOSI-IBM, MOSI-iFlytek datasets. And we also list the results of the SOTA model, Self-MM, on the original MOSI dataset in the last row of the table for the performance comparison between Self-MM in the ideal world and real world. As we can see from the results, Self-MM obtains the best results on the MOSI-Gold dataset than the other datasets, which demonstrates that the ASR errors hurt the MSA models. We also observe that the better ASR model can help the MSA models achieve better performance. But it should be noted that, according to the analysis in the previous section, current ASR models still can not produce satisfactory results for the MSA models in the real world. 

Comparison between the feature-based models including TFN, LMF, and MulT and finetuning-based baselines such as MISA and Self-MM, we can find that finetuning-based models obtain better results. We consider that the finetuning-based models can adapt the BERT encoder to the target task and learning more informative textual representations, which also makes them benefit more as the quality of texts increases. 

Comparing to the baselines especially Self-MM, our model achieves better performance in all evaluation metrics since our model can detect the substitution error of the sentiment words and then refine the word embeddings to reconstruct the sentiment semantics in the textual modality by filtering out useless information from the input words and incorporating useful information from the candidate words generated by the language model. We also observe that the improvement of our model compared with Self-MM on MOSI-iFlytek is smaller. We consider that the main reason is fewer sentiment word substitution errors on MOSI-iFlytek.

\begin{table}[tbp]
	\centering
	\resizebox{\linewidth}{!}{
		\begin{tabular}{l|l|llllll}
			\hline
			\multicolumn{1}{c|}{Models}            & \multicolumn{1}{c}{Has0-Acc $\uparrow$}  &  \multicolumn{1}{c}{Non0-Acc $\uparrow$}  & \multicolumn{1}{c}{MAE $\downarrow$}  \\                           
			
			\hline\hline
			\multicolumn{1}{c|}{SWRM} & \multicolumn{1}{c}{\textbf{74.58}} & \multicolumn{1}{c}{\textbf{75.70}} &  \multicolumn{1}{c}{\textbf{90.56}}   \\
			\multicolumn{1}{c|}{w/o Position} & \multicolumn{1}{c}{73.59}  &  \multicolumn{1}{c}{74.57}    &  \multicolumn{1}{c}{93.67}  \\
			\multicolumn{1}{c|}{w/o Attention} & \multicolumn{1}{c}{74.17}  &  \multicolumn{1}{c}{75.42} &    \multicolumn{1}{c}{91.53}  \\
			\multicolumn{1}{c|}{w/o Multi-modal} & \multicolumn{1}{c}{73.82}  &  \multicolumn{1}{c}{75.09} &  \multicolumn{1}{c}{91.22}\\
			
			\hline
		\end{tabular}
	}
	\caption{Ablation analysis of our proposed model evaluated on the MOSI-SpeechBrain dataset. The best results are in \textbf{bold}.}
	\label{tab:ablation_study}
\end{table}

\begin{figure}[t]
\centering
\includegraphics[width=\columnwidth]{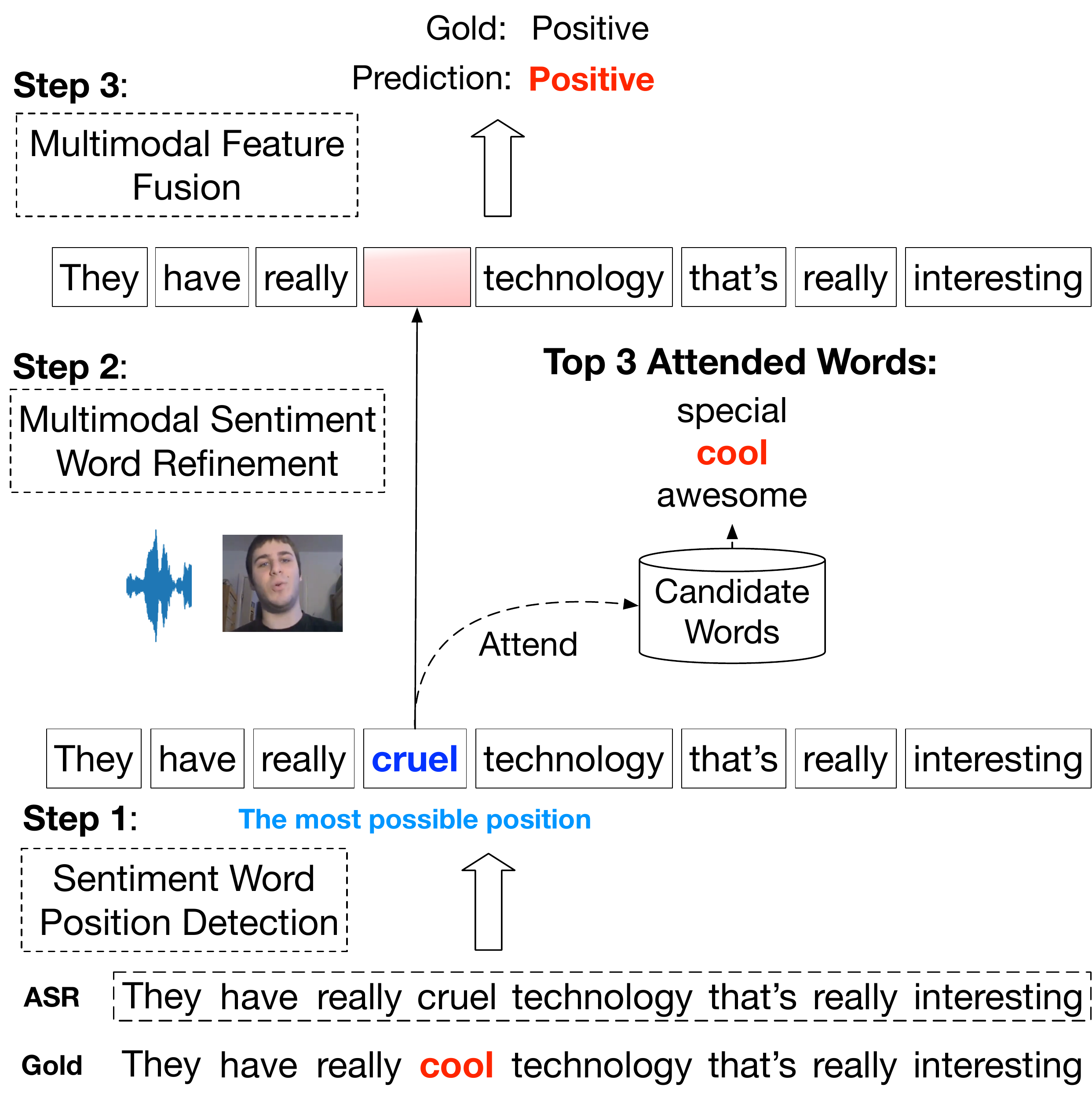} 
\caption{Case study for the SWRM.}.
\label{case}
\end{figure}

\subsection{Ablation Study}
We conduct the ablation experiments to distinguish the contribution of each part. There are several different variants of our model. \textbf{SWRM} is our proposed full model. \textbf{SWRM w/o Position} does not use the sentiment word position location module and only uses the information of the special word [MASK] to dynamically refine all words. \textbf{SWRM w/o Attention} only incorporates the information of the special word [MASK] to refine the word in the multimodal sentiment word refinement module. \textbf{SWRM w/o Multi-modal} only performs the multimodal sentiment word attention and multimodal gating network based on the textual features without the acoustic and visual features. 

Table \ref{tab:ablation_study} shows the results of the variants of our model. After ablating the sentiment word position location module, SWRM w/o Position obtains worse results than SWRM, which indicates that finding the right word for refinement is very important. The comparison between SWRM w/o Attention and SWRM w/o Position further demonstrates this conclusion. SWRM w/o Attention first detects the right position and then incorporates the information of the special word [MASK], which achieves better performance than SWRM w/o Position. But SWRM w/o Attention is still worse than SWRM, which shows using the attention network to incorporating extra information from the candidate words is useful for refinement. Comparing the SWRM w/o Multi-modal between SWRM, we can find that the model benefits from the visual and acoustic features. It is in line with our expectations since the sentiment information provided by the multimodal features can help the model detect the sentiment word and incorporate the sentiment-related information from the candidate words.

\subsection{Case Study}
To have an intuitive understanding of our proposed model, we show a case in Figure \ref{case}. We can see that our model first detects the most possible position based on the context and then finds that the input word in the position may be recognized incorrectly since there is a mismatch between the negative word ``cruel" and either the smile or the excited tone. Hence our model decides to incorporate the related sentiment information from the candidate words to refine the word embedding. As shown in Figure \ref{case}, our model pays more attention to the candidate words "special", "cool", and "awesome". The word "cool" is exactly the gold word and the others have the same sentiment polarity as it. Beneficial from the attended candidate words, our model refines the input word and reconstructs its sentiment semantics. Finally, the refined word embeddings are fed into the multimodal feature fusion module to predict the sentiment label.

\section{Conclusion}
In this paper, we observe an obvious performance drop when the SOTA MSA model is deployed in the real world, and through in-depth analysis, we find that the sentiment word substitution error is a very important factor causing it. To address it, we propose the sentiment word aware multimodal refinement model, which can dynamically refine the word embeddings and reconstruct the corrupted sentiment semantics by incorporating the multimodal sentiment information. We evaluate our model on MOSI-SpeechBrain, MOSI-IBM, and MOSI-iFlytek and the results demonstrate the effectiveness of our approach.  For future work, we will explore leveraging the multimodal information to detect the sentiment word positions. 

\section*{Acknowledgments}
This work was supported by the following Grants: National Natural Science Foundation of China (No. 62176078), National Key R\&D Program of China (No. 2018YFB1005103).

\bibliography{anthology}
\bibliographystyle{acl_natbib}

\appendix

\section{WER Results on MOSI} \label{appdendix_a}

\begin{table}[htp]
	\centering
	\resizebox{0.7\linewidth}{!}{
		\begin{tabular}{llll}
			\hline
			\multicolumn{1}{c}{API} & \multicolumn{1}{c}{SpeechBrain} & \multicolumn{1}{c}{IBM} & \multicolumn{1}{c}{iFlytek} \\
			\hline\hline
			\multicolumn{1}{c}{WER} & \multicolumn{1}{c}{37.54}  & \multicolumn{1}{c}{29.70} & \multicolumn{1}{c}{23.91} \\ 
		    \hline
		\end{tabular}
	}
	\caption{WER results on the MOSI dataset.}
\end{table}

\section{Hyper-parameter Settings} \label{appdendix_b}

\begin{table}[htp]
	\centering
	\resizebox{0.9\linewidth}{!}{
		\begin{tabular}{llll}
			\hline
			\multicolumn{1}{c}{Dataset} & \multicolumn{1}{c}{MOSI-SpeechBrain} & \multicolumn{1}{c}{MOSI-IBM} & \multicolumn{1}{c}{MOSI-iFlytek} \\
			\hline\hline
			\multicolumn{1}{c}{Batch Size} & \multicolumn{1}{c}{64}  & \multicolumn{1}{c}{128} & \multicolumn{1}{c}{64} \\
			\multicolumn{1}{c}{$d_x^l$}  &  \multicolumn{1}{c}{768} &  \multicolumn{1}{c}{768} &  \multicolumn{1}{c}{768}\\
			\multicolumn{1}{c}{$d^{v}_h$} & \multicolumn{1}{c}{16}  & \multicolumn{1}{c}{32} & \multicolumn{1}{c}{16}\\
			\multicolumn{1}{c}{$d^{a}_h$}   & \multicolumn{1}{c}{32}  & \multicolumn{1}{c}{32}& \multicolumn{1}{c}{32}\\
			\multicolumn{1}{c}{$d^{va}_h$} & \multicolumn{1}{c}{48}& \multicolumn{1}{c}{64}& \multicolumn{1}{c}{48}\\
			
			\multicolumn{1}{c}{$d_v^l$} & \multicolumn{1}{c}{32} & \multicolumn{1}{c}{64}& \multicolumn{1}{c}{32}\\
			
			\multicolumn{1}{c}{$d_v^a$}& \multicolumn{1}{c}{16}& \multicolumn{1}{c}{32}& \multicolumn{1}{c}{16}\\
			\multicolumn{1}{c}{$d_v^v$}  & \multicolumn{1}{c}{32}& \multicolumn{1}{c}{16}& \multicolumn{1}{c}{32}\\
			\multicolumn{1}{c}{$d_v^f$} & \multicolumn{1}{c}{128} & \multicolumn{1}{c}{64}   & \multicolumn{1}{c}{128} \\
		
			\hline
		\end{tabular}
	}
	\caption{The hyper-parameters used in training for the three datasets.}
\end{table}

\end{document}